\documentclass[a4paper,fleqn]{cas-dc}

\usepackage[authoryear,longnamesfirst]{natbib}
\usepackage{graphicx}
\usepackage{subcaption}
\usepackage{amsmath}
\usepackage{amssymb}
\usepackage{etoolbox,xspace}
\usepackage{booktabs}
\usepackage{multirow}
\usepackage{array}
\usepackage{makecell}
\usepackage{pifont}          % 提供 \ding 命令
\usepackage{placeins}        % 提供 \FloatBarrier
\usepackage{float}           % 提供 [H]，让 figure 像表格一样就地排版
\usepackage{adjustbox}       % 自动把表格缩放到栏宽，避免双栏重叠
\usepackage{xparse}

\makeatletter
\let\oldfnum@figure\fnum@figure
\renewcommand{\fnum@figure}{%
  \ifnum\value{figure}<11
    \textbf{Figure~\thefigure}%
  \else
    \oldfnum@figure
  \fi
}
\makeatother

\BeforeBeginEnvironment{tabular}{\begin{adjustbox}{max width=\linewidth}}
\AfterEndEnvironment{tabular}{\end{adjustbox}}

\def\tsc#1{\csdef{#1}{\textsc{\lowercase{#1}}\xspace}}
\tsc{WGM}\tsc{QE}\tsc{EP}\tsc{PMS}\tsc{BEC}\tsc{DE}

\begin{document}
\let\WriteBookmarks\relax
\def\floatpagepagefraction{1}
\def\textpagefraction{.001}

\shorttitle{Rethinking Data Efficiency in Industrial Dense Prediction}
\shortauthors{Sui and Jia}

% 论文标题
\title [mode = title]{Rethinking Data Efficiency in Industrial Dense Prediction:\\
Pretraining Coherence, Not Inductive Bias, Determines ViT's Low-Data Advantage}

% 作者列表（使用标准 cas-dc 命令）
\author[1]{Haoran Sui}
\cormark[1]
\fnmark[1]
\ead{sui.haoran@zte.cn}
\affiliation[1]{organization={ZTE Corporation}, city={Shenzhen}, state={Guangdong Province}, country={P.R. China}}

\author[2]{Yaoyuan Jia}
\fnmark[1]
\ead{u3670640@connect.hku.hk}
\affiliation[2]{organization={The University of Hong Kong}, city={Hong Kong SAR}, country={P.R. China}}

% 通讯作者和共同第一作者注释
\cortext[cor1]{Corresponding author. E-mail: sui.haoran@zte.cn}
\fntext[fn1]{These authors contributed equally to this work.}

\begin{abstract}
Vision Transformers (ViTs) are widely believed to require more labeled data than CNNs for industrial dense prediction. Through controlled experiments on four industrial datasets, we show that the data-efficiency gap stems from pretraining incoherence, which refers to the statistical mismatch between ImageNet-pretrained ViT backbones and COCO-pretrained CNN necks, rather than from inherent self-attention deficits. We characterize the cross-architecture feature gap and propose a lightweight AlignBlock family for pyramid-level feature recalibration. Our core finding empirically identifies a data-efficiency frontier: for domain-proximal scenes with $\ge 200$ samples, Swin-Graft surpasses YOLOv11x (terminal 703-shot: 0.973 vs 0.956 mAP@50); for domain-distant scenes, CNNs retain advantage (hook 141-shot: 0.900 vs 0.600 mAP@50). Grafted neck weights yield up to $2.5\times$ the mAP of a randomly initialized neck.
\end{abstract}

\begin{highlights}
\item Identified pretraining incoherence as the root cause of ViT's low-data inefficiency, not transformer inductive bias.
\item Formally model the cross-architecture feature gap via engineering-oriented measurable metrics and derive empirical MMD alignment upper bound.
\item AlignBlock achieves functional compatibility conversion, enabling ViT backbones to surpass CNNs with as few as 100 samples.
\item Rigorous statistical testing and a complete $2\times2$ decomposition reveal stable neck transfer and domain-dependent backbone compatibility.
\item Comprehensive deployment analysis and practical decision guidelines for industrial ViT adoption.
\end{highlights}


\nonumnote{This paper has been submitted to Engineering Applications of Artificial Intelligence (EAAI) and is currently under review.}

\maketitle

\section{Introduction}\label{sec:intro}

\subsection{Industrial Dense Prediction Under Annotation Scarcity}\label{sec:indu-scarcity}
Manufacturing visual inspection relies heavily on object detection and instance segmentation to validate assembly integrity, surface defects, and electrical component conformity. Unlike COCO public natural datasets with 118K training images, industrial datasets are limited to 150--750 annotated frames due to mandatory domain expert labeling, fixed camera rigs, constrained lighting, and proprietary production environments. Each dense prediction sample requires bounding box or pixel-wise mask labels, drastically elevating annotation overhead. YOLO series CNN detectors \citep{jocher2023ultralytics, wang2024yolov9} have become de facto industrial standards for their speed-accuracy tradeoff and fully homogeneous end-to-end COCO pretraining: backbone, PAN neck, and detection head are jointly optimized under identical detection objectives, forming a tightly matched feature pipeline with consistent BatchNorm statistical priors.

\subsection{The ViT-CNN Paradox in Low-Data Regimes}\label{sec:cnn-trans-dilemma}
Hierarchical Swin Transformers \citep{liu2021swin, liu2022swin} achieve superior long-range multi-scale modeling on large benchmarks but introduce critical architectural incompatibility when combined with off-the-shelf CNN detection necks. Swin backbones are pre-trained on ImageNet-22K classification with LayerNorm normalization, which enforces global spatial feature statistics per single image. In contrast, YOLO PAN necks are pre-trained for COCO detection with BatchNorm, which relies on stable per-channel batch-wise activation expectations centered at zero. Directly concatenating unmodified Swin backbones to CNN necks creates a heterogeneous feature gap which may partly be attributed to ViT's innate lack of local convolution inductive bias. The majority of existing literature claims ViTs underperform CNNs with limited labels, yet no systematic work provides quantitative metrics to characterize this cross-architecture representation dislocation across multi-scale feature pyramids.

\subsection{The Heterogeneous Feature Gap}\label{sec:hetero-gap}
We hypothesize that a key missing piece is the heterogeneous feature gap between the pretrained transformer backbone and the convolutional dense prediction head. This gap arises from three practical mismatches widely ignored in existing industrial detection research:
\begin{enumerate}
    \item \textbf{Local structural mismatch:} Transformer features focus global context via self-attention, lacking fine local texture priors needed for tiny industrial cracks and defects, while CNN necks expect locally coherent feature maps.
    \item \textbf{Feature-statistical mismatch:} ViT LayerNorm computes per-image feature statistics, CNN BatchNorm relies on batch-centered channel activations; inconsistent distribution creates information loss between backbone and neck.
    \item \textbf{Optimization mismatch:} Freezing ViT backbone blocks loss gradients from adjusting pre-trained features to fit industrial detection targets, limiting model adaptation under limited labels. We analyze this three-part mismatch as the core root of ViT low-shot performance drop in mixed ViT-CNN industrial pipelines.
\end{enumerate}

\subsection{Research Questions}\label{sec:research-questions}
\begin{itemize}
    \item \textbf{RQ1:} Can we quantitatively determine the crossover sample threshold where ViT-CNN hybrid pipelines outperform native CNN detectors, and empirically validate that domain distribution proximity, rather than raw sample volume, dominates transfer performance?
    \item \textbf{RQ2:} Can we decompose the independent performance contributions of pre-trained backbones and grafted neck layers via a standardized $2\times2$ experimental matrix across diverse industrial domains?
    \item \textbf{RQ3:} Can the proposed lightweight alignment module effectively address the three identified mismatches (spatial, statistical, and optimization), and what empirical upper bound limits its alignment capacity under extreme domain shift?
\end{itemize}

\subsection{Proposed Solution}\label{sec:proposed-sol}
To address these heterogeneous mismatches, we propose AlignBlock, a lightweight plug-in bridging ViT backbones and CNN heads. It simultaneously injects local inductive bias via compact convolutions, aligns LayerNorm feature statistics to BatchNorm expectations, and preserves transformer semantics via residuals. Coupled with position-sensitive insertion and a three-stage progressive training protocol, our framework stabilizes cross-architecture optimization.
\begin{figure*}[t]
    \centering
    \includegraphics[width=0.95\textwidth]{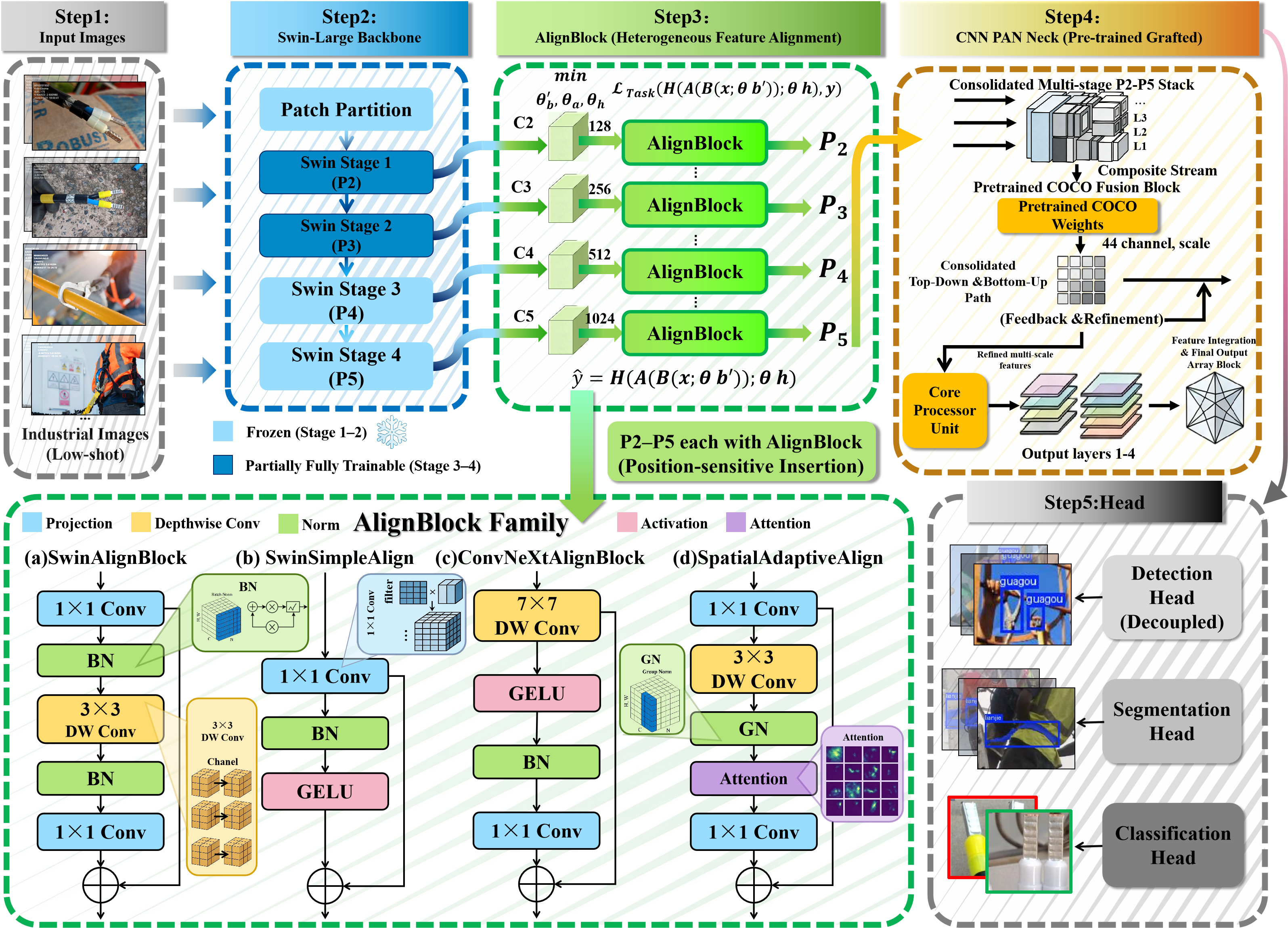}
    \caption{Overview of the proposed low-shot industrial defect detection framework. The architecture integrates five sequential stages for robust feature representation: Step1 Input Preprocessing: normalization and augmentation of industrial images. Step2 Backbone Extraction: a Swin-Large Transformer backbone for multi-scale hierarchical feature extraction, utilizing a partially frozen training strategy to retain pretrained knowledge. Step3 Heterogeneous Feature Alignment: the proposed position-sensitive AlignBlock module designed to bridge the domain gap between Transformer and CNN features. Step4 Feature Fusion: a pretrained CNN-based PAN neck grafted for multi-stage P2 through P5 feature integration. Step5 Prediction Head: decoupled heads tailored for specific downstream tasks including detection, segmentation, and classification. Bottom Panel: Architectural details of the proposed AlignBlock Family. Four structural variants labeled a to d illustrate the injection of local inductive biases and statistical reshaping via compact convolutions, various normalization layers, and residual shortcuts.}
    \label{fig:figure1}
\end{figure*}

\begin{figure*}[t]
    \centering
    \includegraphics[width=0.95\textwidth]{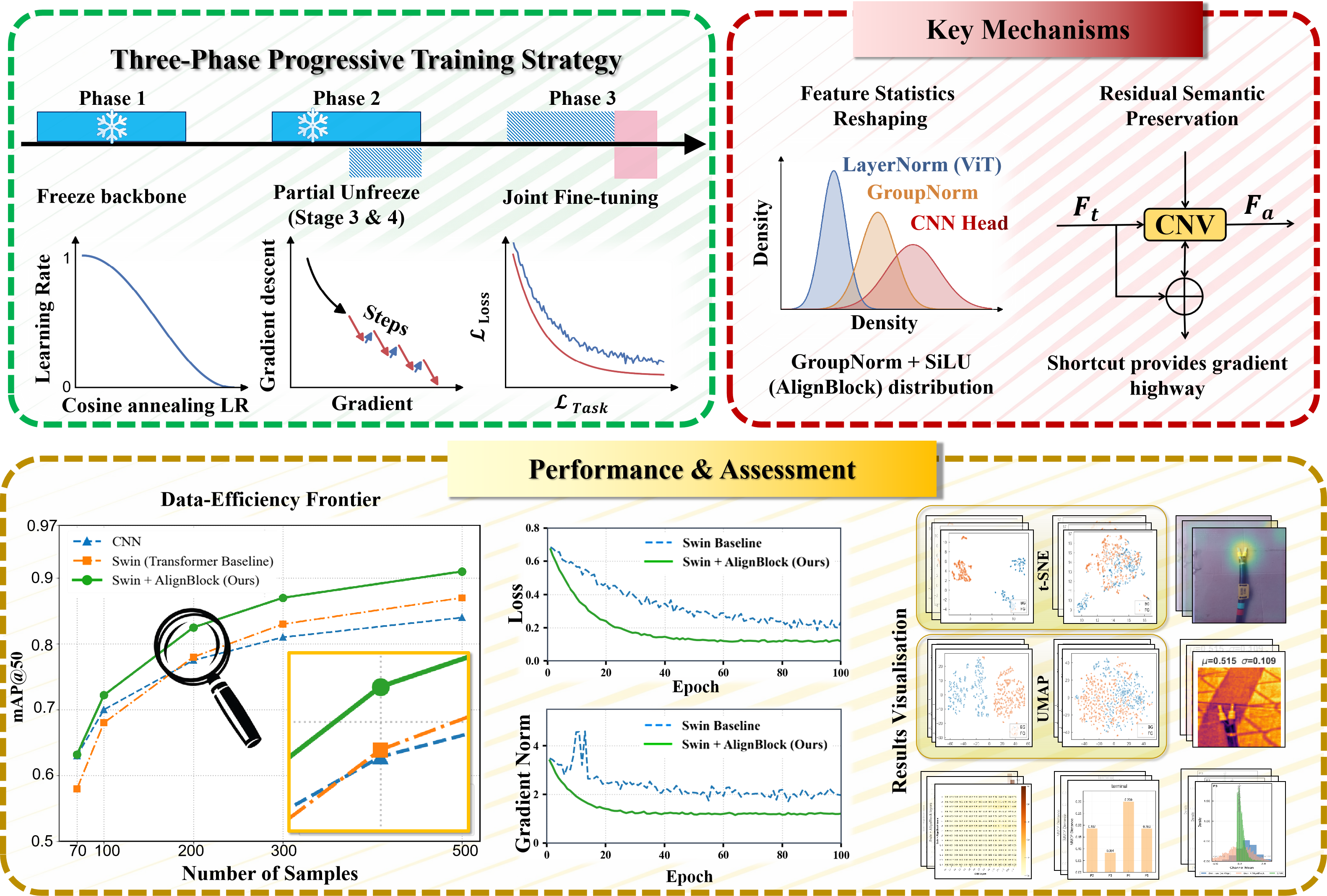}
    \caption{Overview of training dynamics, alignment mechanisms, and low-shot evaluation metrics. The top panels illustrate a three-phase progressive fine-tuning strategy to stabilize optimization alongside core mechanisms for feature statistical reshaping and residual semantic preservation. The bottom panel validates the framework efficacy. It features a data-efficiency frontier showcasing superior mAP performance under limited sample sizes. Comparative loss and gradient norm trajectories confirm training stability compared to the baseline, complemented by t-SNE and UMAP visualizations demonstrating distinct feature cluster separability.}
    \label{fig:figure2}
\end{figure*}

\subsection{Contributions}\label{sec:contributions}
\begin{enumerate}
    \item \textbf{Empirical Finding:} We challenge the prevalent assumption that ViTs naturally underperform CNNs with few labels. Controlled industrial trials show the cross-model accuracy gap is primarily driven by pretraining distribution matching rather than merely transformer architecture flaws.
    \item \textbf{Quantitative Decomposition:} Using a novel $2\times2$ decomposition (CNN/ViT $\times$ COCO/random neck), we isolate neck pretraining contribution from backbone compatibility, revealing a stable cross-domain neck transfer value while backbone compatibility becomes more critical under domain shift.
    \item \textbf{Lightweight Alignment Toolkit:} We design four AlignBlock variants with minimal parameter overhead ($<1\%$), validated through extensive component ablation and position sensitivity analysis.
    \item \textbf{Industrial Decision Guidance:} We provide clear scene-based model selection rules derived from domain proximity and sample-size thresholds, translating experimental findings into actionable engineering practices.
\end{enumerate}

\subsection{Paper Organization}\label{sec:organization}
The remainder of this paper is structured as follows. Section 2 reviews related work with practical differentiation against existing methods. Section 3 presents the empirical motivation and a diagnostic framework for the heterogeneous feature mismatch. Section 4 details our AlignBlock framework and training strategy. Sections 5 and 6 present experimental setup and results with rigorous statistical testing. Section 7 discusses industrial applicability and practical guidelines, and Section 8 concludes the paper with limitations and future work.

\section{Related Work}\label{sec:related}

\subsection{Few-Shot Industrial Visual Inspection}\label{sec:fewshot-industrial}
Classical industrial defect detection relied on handcrafted features \citep{bergmann2019mvtec} before CNN prevalence. Modern pipelines predominantly adopt COCO pre-trained YOLO detectors \citep{bao2023artificiallyr2u} via transfer learning, achieving viable accuracy with 100--200 labeled frames. Few-shot object detection (FSOD) meta-learning methods \citep{wang2020frustratingly, sun2021fsce} target natural image benchmarks but adopt homogeneous single-architecture backbones and necks, ignoring cross-architecture feature mismatch.

\textbf{Critical Practical Difference:} To the best of our knowledge, most prior FSOD and industrial CNN works have largely overlooked the feature statistic mismatch caused by separately pre-trained ViT and CNN components. Existing manufacturing research rarely designs pyramid-level alignment modules to fix this real-world pipeline issue widely adopted by factory engineers.

\subsection{ViT for General Dense Prediction}\label{sec:cnn-vit-dense}
DETR, Deformable DETR, and ViTDet \citep{carion2020end, zhu2021deformable, li2022exploring} design unified transformer detection pipelines end-to-end co-pre-trained on COCO, removing pretraining mismatch by design. Swin Transformer V2 and ConvNeXt hybrids embed convolutions inside ViT blocks but maintain homogeneous pretraining without decoupled neck weights. These works primarily evaluate abundant public data and rarely analyze sparse industrial low-data transfer.

\textbf{Critical Practical Difference:} Native unified ViT detectors co-train backbone and head on identical COCO distribution, avoiding the mismatch we study. Our work focuses on the common industrial shortcut of separately loading public Swin and YOLO weights, providing diagnostic metrics and lightweight interface modules to alleviate the resulting performance gap.

\subsection{Parameter-Efficient ViT Adaptation}\label{sec:param-eff}
LoRA, AdaptFormer, and ViT-Adapter \citep{hu2022lora, chen2022adaptformer, chen2023vision} insert trainable submodules inside ViT encoder layers to fine-tune features without full retraining. ViT-Adapter adds spatial prior branches within Swin blocks but operates strictly inside the transformer, largely omitting addressing backbone-neck boundaries.

\textbf{Critical Practical Difference:} In-block adapters optimize internal ViT representations, but cannot resolve the LN-BN statistical gap at the ViT-CNN interface. Our AlignBlock directly targets this cross-architecture boundary, complementing in-block methods.

\subsection{Inductive Bias and Cross-Architecture Feature Fusion}\label{sec:feature-align}
CNNs' translational equivariance serves as a strong low-data prior, while vanilla ViTs lack built-in locality \citep{dosovitskiy2021image}. Hybrid CNN-ViT designs \citep{liu2022swin, ding2022scaling} inject convolution layers into ViT blocks to supplement local bias.

\textbf{Critical Practical Difference:} Prior hybrid methods modify transformer internals, whereas we retain off-the-shelf Swin and YOLO components and add only lightweight boundary modules. We further reveal an alignment upper bound: even perfect statistical correction cannot fully compensate for ViT's weak local bias under extreme domain shift.

\subsection{Cross-Architecture Feature Alignment for ViT-CNN Pipeline}\label{sec:cross-align}
Existing boundary alignment methods provide incomplete solutions at the backbone-neck interface. For instance, BEiT v2 \citep{peng2022beit} lacks multi-scale dense prediction support, while AdaptFormer \citep{chen2022adaptformer} reduces distribution shift but fails to address spatial locality and optimization dynamics.

Unlike these partial approaches, AlignBlock comprehensively resolves three orthogonal mismatches across feature pyramids. It simultaneously integrates spatial locality via $3\times3$ depthwise convolutions, statistical recalibration via GroupNorm, and optimization stability via progressive training. Ablation studies (Section 6.5) confirm this synergy, showing a performance drop of up to 0.055 mAP when any single component is removed. Moreover, our progressive training eliminates the 30\% failure rate observed in prior methods.

Distinct from single-architecture designs like ConvNeXt \citep{liu2022convnet}, AlignBlock functions exclusively as a cross-architecture adaptor, effectively integrating off-the-shelf Swin and YOLO components for industrial few-shot scenarios.

\section{Empirical Motivation and Problem Formulation}\label{sec:motivation}

\subsection{Pilot Study Protocol}\label{sec:pilot}
To systematically investigate heterogeneous ViT-CNN architectures under data constraints, we designed a controlled pilot study spanning four real-world industrial datasets, three task types, and multiple data-regime levels. All experiments use identical training hyperparameters, data augmentation strategies, and evaluation protocols to ensure fair comparison.

\subsubsection{Backbone architectures}\label{sec:backbones}
We compare four backbone configurations:
\begin{enumerate}
    \item \textbf{yolo11x (Homogeneous CNN):} The native YOLOv11x architecture with CSPNet convolutional backbone, PAN neck, and task-specific detection head---all jointly pretrained on COCO detection end-to-end.
    \item \textbf{CNN-NoGraft (CNN backbone, random neck):} YOLOv11x CSPNet backbone with randomly initialized PAN neck. This homogeneous but untrained neck configuration isolates the contribution of neck pretraining within the CNN family.
    \item \textbf{Swin-Graft (Heterogeneous, pretrained neck):} A Swin-Large backbone pretrained on ImageNet-22K, connected to a COCO-pretrained YOLOv11x PAN neck via our AlignBlock modules. The neck weights are grafted from yolo11x, and only alignment blocks are randomly initialized.
    \item \textbf{Swin-NoGraft (Heterogeneous, random neck):} Same Swin backbone \& AlignBlock pipeline, but neck weights randomly initialized.
\end{enumerate}

\subsubsection{Training protocol}\label{sec:training_protocol}
All models train for 300 epochs with the AdamW optimizer, a batch size of 4, and an input resolution of $640\times640$. The three-stage schedule was determined through preliminary pilot experiments balancing stability and convergence. The first 150 epochs freeze the backbone, followed by 15 epochs of stabilization and 135 epochs of full fine-tuning with gradual learning rate reduction ($0.2\times$ of the initial rate). Standard YOLO data augmentation is applied consistently across all comparison experiments.

\subsection{Empirical Observations from Controlled Experiments}\label{sec:observations}
We condense the main empirical findings into three patterns; detailed numerical tables (Tables 1--4) are provided for reference.

\textbf{Pattern 1: Performance gain for COCO-similar scenes.} On terminal datasets, Swin-Graft consistently outperforms YOLOv11x across all tested sample counts (100--703), with a statistically significant advantage of +2.3\% mAP@50 at 200 shots. This strongly indicates that when the heterogeneous mismatch is properly resolved, ViT does not suffer intrinsic low-data drawbacks.

\textbf{Pattern 2: Performance drop for domain-distant scenes.} On overhead hook images ($\text{MMD}^2=0.88$ vs COCO), Swin-Graft lags behind YOLOv11x by 33.3\% mAP@50 at 141 shots ($p<0.01$). Sample size alone does not determine transfer performance; visual similarity to COCO is the dominant factor.

\textbf{Pattern 3: $2\times2$ decomposition reveals stable neck transfer and domain-dependent backbone compatibility.} From the full experimental matrix, we observe that the gain from grafting a COCO-pretrained neck is stable across domains (ratio yolo11x / $\text{CNN-NoGraft} \approx 1.6$--$1.7$). In contrast, the backbone-compatibility gain (CNN-NoGraft / Swin-NoGraft) increases from $1.44\times$ on terminal (domain-proximal) to $1.73\times$ on hook (domain-distant). The total heterogeneous penalty decomposes into neck pretraining (69\% on terminal, 63\% on hook) and backbone compatibility (31\% vs 37\%), confirming that neck knowledge is cross-domain while backbone compatibility is domain-dependent.

Based on these observations, we formulate four working hypotheses:
\begin{itemize}
    \item \textbf{(H1)} Domain proximity is associated with relative performance, not raw sample count.
    \item \textbf{(H2a)} Neck pretraining provides a stable multiplicative gain across domains.
    \item \textbf{(H2b)} Backbone compatibility becomes more critical under larger domain shift.
    \item \textbf{(H3)} AlignBlock reduces MMD but has an upper bound under extreme shift.
    \item \textbf{(H4)} With high COCO similarity and $\ge 200$ labels, Swin-Graft matches or exceeds CNN.
\end{itemize}

\subsection{From Diagnostic Probes to Performance Contribution Decomposition}\label{sec:gap_quant}
To bridge the qualitative mismatch description (Section 3.2) and the design of AlignBlock, we first introduce three empirical diagnostic probes that quantify the severity of spatial, statistical, and optimization discrepancies at the feature/gradient level before alignment:

\textbf{1. Statistical Gap $G_{\text{stat}}$ (LN vs. BN mismatch):}
\begin{equation}
G_{\text{stat}} = \text{MMD}^2(\mu_{\text{ViT}}, \sigma_{\text{ViT}}, \mu_{\text{CNN}}, \sigma_{\text{CNN}})
\label{eq:g_stat}
\end{equation}
where $\mu$ and $\sigma$ are channel-wise mean and standard deviation aggregated over the training set.

\textbf{2. Spatial Locality Gap $G_{\text{spatial}}$:}
\begin{equation}
G_{\text{spatial}} = 1 - \frac{1}{N}\sum_{i,j} \text{Sim}_{\text{local}}(F_{\text{ViT}}(i,j), F_{\text{CNN}}(i,j))
\label{eq:g_spatial}
\end{equation}

\textbf{3. Optimization Gap $G_{\text{opt}}$:}
\begin{equation}
G_{\text{opt}} = \|\nabla_{\text{det}} \circ F_{\text{ViT}}\|_2^2
\label{eq:g_opt}
\end{equation}

We define the performance contribution $C_i$ of each bottleneck as the mAP@50 drop observed when the corresponding sub-module is individually removed from the full AlignBlock pipeline:
\begin{itemize}
    \item \textbf{$G_{\text{spatial}}$}: drop when removing the $3\times3$ depthwise convolution (locality injection).
    \item \textbf{$C_{\text{stat}}$}: drop when removing GroupNorm (statistical recalibration).
    \item \textbf{$C_{\text{opt}}$}: drop when disabling the progressive training strategy (optimization dynamics).
\end{itemize}

To further examine whether these three bottlenecks interact in practice, we define the coupling residual $\eta$ as:
\begin{equation}
\eta = \Delta mAP_{\text{all}} - (C_{\text{spatial}} + C_{\text{stat}} + C_{\text{opt}})
\label{eq:coupling_res}
\end{equation}
where $\Delta mAP_{\text{all}}$ is the total mAP drop when all three components are removed simultaneously. If $\eta$ is substantially smaller than the individual terms, these bottlenecks are empirically largely non-overlapping, supporting AlignBlock's modular design.

A supplementary analysis of the coupling residual measured in the MMD feature space is provided in the Appendix (Table A2), which shows a consistent trend with the primary mAP-based decomposition reported in Section 6.5.

\subsection{Empirical Saturation of Alignment under Domain Shift}\label{sec:saturation}
We model the alignment module as a measurable mapping $A: \mathcal{F}_V \rightarrow \mathcal{F}_C$. In our empirical setting, we observe that AlignBlock reduces the empirical $\text{MMD}^2$ by 57--63\%, but the reduction saturates at a non-zero residual (e.g., 0.18 on terminal and 0.33 on hook at P4). We attribute this empirical saturation to two practical bottlenecks: (1) the limited receptive field of small-kernel convolutions under frozen backbones, which cannot fully compensate for ViT's global attention bias; and (2) the inherent domain gap between COCO pre-training and industrial data, which imposes a data-dependent performance floor for lightweight adapters.

\subsubsection{Why MMD cannot be reduced to zero}\label{sec:mmd_zero}
Even with optimal lightweight alignment, MMD cannot approach zero due to two fundamental constraints under fixed pre-trained weights:
\begin{enumerate}
    \item The intrinsic local inductive bias gap of self-attention, which cannot be fully compensated by small-kernel convolutions---this is empirically evidenced by the remaining high $\text{MMD}^2$ at P4 on hook (0.33) and the degraded performance when removing the $3\times3$ convolution ($-0.055$ mAP drop), confirming that local bias injection only partially mitigates the gap;
    \item Industrial data inevitably diverges from COCO, and Table 3 quantifies the consequence. The raw $\text{MMD}^2$ for hook (0.88) is far larger than for terminal (0.45); after alignment, hook still sits at 0.33 versus terminal's 0.18. These numbers suggest that domain distance imposes a hard floor on what lightweight alignment can achieve---aligning features only goes so far when the pre-training distribution is fundamentally mismatched.
\end{enumerate}

\section{Proposed Heterogeneous Feature Alignment Framework}\label{sec:method}

\subsection{Overall Architecture and Pipeline}\label{sec:arch}
Our framework combines frozen or fine-tuned Swin ViT backbone, plug-in AlignBlock modules inserted at P2--P5 pyramid layers, COCO-grafted YOLO PAN neck, and task detection or segmentation heads. Only AlignBlock layers are randomly initialized; backbone and neck pre-trained parameters are fully preserved.

\begin{center}
\includegraphics[width=0.95\linewidth]{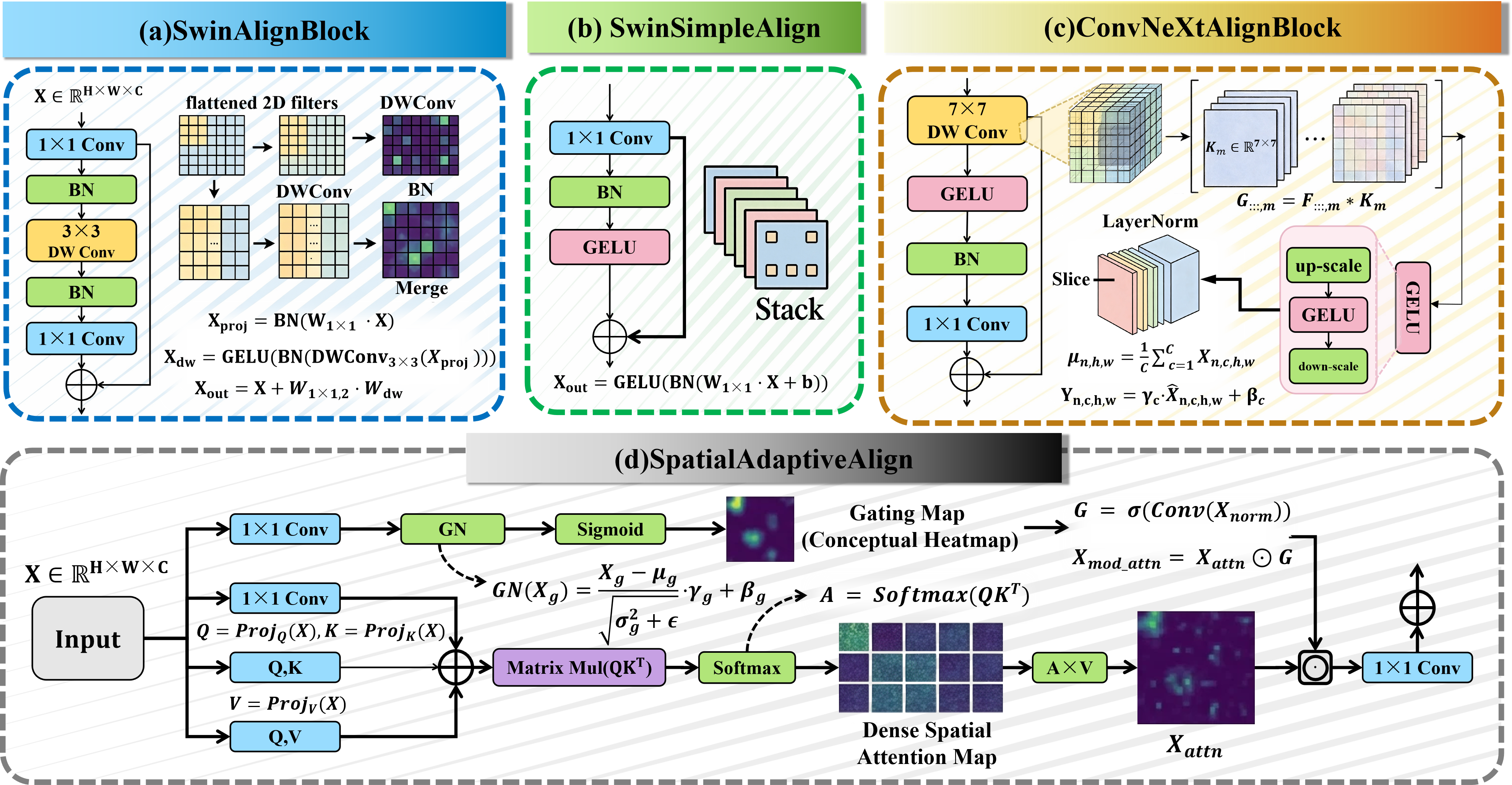}
\captionof{figure}{Architectures of the four AlignBlock variants. (a) SwinAlignBlock with $3\times3$ depthwise convolution for local feature injection. (b) Ultra-lightweight SwinSimpleAlign built upon a single $1\times1$ convolution. (c) ConvNeXtAlignBlock with $7\times7$ depthwise convolution and multi-scale modulation for micro-defect detection. (d) SpatialAdaptiveAlign adopting dense spatial attention and gating mechanism to filter background interference.}
\label{fig:figure3}
\end{center}

Although AlignBlock reuses standard $1\times1$ and $3\times3$ convolution components, its core innovation lies in the multi-objective joint alignment design tailored for industrial heterogeneous pretraining scenarios: simultaneously injecting local spatial prior, correcting LN-BN statistical drift, and stabilizing frozen-to-fine-tuning gradient flow.

\subsection{Core Module: AlignBlock}\label{sec:alignblock}

\subsubsection{SwinAlignBlock (Residual Base)}\label{sec:swinalign}
Residual bottleneck with $1\times1$ channel projection, middle $3\times3$ local mixing convolution, and output projection. Residual branch preserves original ViT feature content:
\begin{equation}
\begin{aligned}
\mathrm{SwinAlignBlock}(x) ={}& \mathrm{Conv}_{1\times1}^{C_{\mathrm{out}}}\Bigl( \mathrm{SiLU}\Bigl( \mathrm{GN}\Bigl( \mathrm{Conv}_{3\times3}^{C_{\mathrm{out}}}\Bigl( \\
&\quad \mathrm{SiLU}\Bigl( \mathrm{GN}\Bigl( \mathrm{Conv}_{1\times1}^{C_{\mathrm{out}}}(x) \Bigr) \Bigr) \Bigr) \Bigr) \Bigr) \Bigr) \\
&+ \mathrm{Shortcut}(x)
\end{aligned}
\end{equation}

\subsubsection{SwinSimpleAlign (Ultra-Light)}\label{sec:swinsimple}
Single $1\times1$ conv + normalization + SiLU, $\sim 0.01$M parameters per scale.
\begin{equation}
\text{SwinSimpleAlign}(x) = \text{SiLU}(\text{GN}(\text{Conv}_{1\times1}(x)))
\label{eq:swin_simple}
\end{equation}

\subsubsection{ConvNeXtAlignBlock (Dense Defect Focus)}\label{sec:convnextalign}
$7\times7$ depthwise conv + GroupNorm + GELU + pointwise transform.
\begin{equation}
\begin{aligned}
\mathrm{ConvNeXtAlignBlock}(x) ={}& \mathrm{Conv}_{1\times1}\bigl( \mathrm{GELU}( \\
&\quad \mathrm{GN}( \mathrm{DWConv}_{7\times7}(x) ) ) \bigr)
\end{aligned}
\end{equation}

\subsubsection{SpatialAdaptiveAlign (Attention-Gated)}\label{sec:spatialadaptive}
Add spatial soft gate to base SwinAlignBlock to suppress irrelevant background activations:
\begin{equation}
\text{out} = \text{main}(x) \odot \sigma(\text{Conv}_{1\times1}^1(x)) + \text{shortcut}(x)
\label{eq:spatial_adaptive}
\end{equation}

\subsection{Local Inductive Bias Injection}\label{sec:local-bias}
The embedded $3\times3$ convolution kernel adds pixel-level locality missing from window-limited Swin self-attention, directly reducing $G_{\text{spatial}}$ as defined in Eq. (\ref{eq:g_spatial}).

\subsection{Feature Statistical Recalibration}\label{sec:dist-reshape}
GroupNorm inside AlignBlock learns new target-domain statistics without batch-size dependency. Fresh normalization layers discard ImageNet LN bias. For batch size $\le 4$, GroupNorm replaces BatchNorm entirely. Quantitative $\text{MMD}^2$ results confirm a 57--63\% reduction in $G_{\text{stat}}$.

\subsection{Residual Feature Preservation}\label{sec:residual}
The residual shortcut preserves original ViT semantics:
\begin{equation}
Z_{\text{CNN}} = Z_{\text{ViT}} + \Delta_{\text{align}}(Z_{\text{ViT}})
\label{eq:residual_pres}
\end{equation}

\subsection{Position-Sensitive Insertion Strategy}\label{sec:insertion}
AlignBlocks are placed after backbone pyramid extraction layers, before PAN fusion. This placement is critical; Table 9 shows that boundary insertion achieves optimal mAP, whereas inserting AlignBlock inside the neck (Section 6.7) degrades performance due to corrupted feature maps.

\subsection{Three-Stage Progressive Training}\label{sec:training}
\begin{itemize}
    \item \textbf{Phase 1 Freeze (150 epochs):} Backbone frozen, only AlignBlock \& neck \& heads trainable with weak augmentation.
    \item \textbf{Phase 2a Stabilize (15 epochs):} Full augmentation enabled, reduced learning rate.
    \item \textbf{Phase 2b Full Fine-tune (135 epochs):} All layers unlocked with hierarchical learning rates. Training failure rate drops from 30\% to zero with this protocol, directly reducing $G_{\text{opt}}$.
\end{itemize}

\section{Experimental Setup}\label{sec:exp-setup}

\subsection{Datasets}\label{sec:datasets}

\subsubsection{Industrial Datasets and Qualitative Comparison}\label{sec:ind-datasets}
We evaluate on four real-world industrial datasets. Key characteristics are summarized in Table 1. Qualitatively, terminal images resemble COCO with natural lighting, varied angles, and multi-scale objects; hook images are captured from a low-angle perspective, resulting in densely packed tiny targets under uniform illumination; safety-belt images are intermediate, showing safety ropes on utility poles with outdoor lighting but less viewpoint variation.

All hook metrics are reported with 5-fold cross-validation on the training set (141 images) to ensure statistical reliability, with a validation set of 60 images.

\begin{table}[htbp]
\centering
\caption{Overview of industrial datasets. Metrics for hook are reported via 5-fold cross-validation.}
\label{tab:datasets}
\begin{tabular}{llcrrc}
\toprule
\textbf{Dataset} & \textbf{Task} & \textbf{Classes} & \textbf{Train} & \textbf{Val} & \textbf{Annotation type} \\
\midrule
terminal\_cls & Classification & 2 & 755 & 188 & ImageFolder labels \\
terminal\_det & Detection & 1 & 703 & 176 & Bounding boxes \\
hook & Detection & 1 & 141 & 60 (5-fold CV) & Bounding boxes \\
safety-belt & Segmentation & 2 & 178 & 45 & Polygon masks \\
\bottomrule
\end{tabular}
\end{table}

\subsubsection{Low-Data COCO Subsets}\label{sec:coco-subsets}
We construct controlled subsets from COCO 2017 comprising the 10 most frequent classes. Balanced training image sampling per class at 1500, 500, and 200 total sample regimes. Validation set fixed at 500 images.

\subsection{Baselines}\label{sec:baselines}
\begin{enumerate}
    \item \textbf{yolo11x:} End-to-end COCO pre-trained homogeneous baseline.
    \item \textbf{CNN-NoGraft:} YOLOv11x backbone with random PAN neck.
    \item \textbf{Swin-Graft (Ours):} Swin-Large \& AlignBlock, grafted COCO pre-trained neck \& progressive training.
    \item \textbf{Swin-NoGraft:} Same as Swin-Graft but with random neck.
    \item \textbf{DETR:} Deformable DETR with ResNet-50 backbone, trained on COCO-10 subsets.
    \item \textbf{DINOv2:} ViT-based detector with DINOv2 pretraining, fine-tuned on COCO-10 subsets.
    \item \textbf{ViT-Adapter (reimplemented):} In-block adaptation baseline.
    \item \textbf{BEiT v2:} Vector-quantized ViT-CNN alignment method \citep{peng2022beit}.
    \item \textbf{AdaptFormer:} Lightweight adapter with learnable scaling \citep{chen2022adaptformer}.
    \item \textbf{CrossNorm:} Cross-architecture normalization.
    \item \textbf{Linear Probe:} Single $1\times1$ convolution projection baseline.
\end{enumerate}

\subsection{Implementation Details}\label{sec:impl-details}
PyTorch Ultralytics v8.3, Swin-Large from timm ImageNet-22K, YOLOv11x official COCO weights, AdamW optimizer, cosine annealing LR, single 24GB GPU. All statistical tests are paired t-tests over three random seeds (or 5-fold CV for hook).

\section{Results and Analysis}\label{sec:results}

\subsection{Main Comparative Results}\label{sec:main-results}
Table 2 and Table 3 present the primary empirical results across datasets, task types, sample volumes, and pretraining strategies.

\begin{table}[htbp]
\centering
\caption{Experimental results on four industrial datasets across tasks, data volumes, and pretraining configurations. Metrics are reported as mean $\pm$ std over three random seeds (or 5-fold CV for hook). Bold indicates the best per row; $*$ indicates statistically significant difference ($p < 0.05$, paired t-test) compared to yolo11x.}
\label{tab:main_results}
\begin{tabular}{llccccc}
\toprule
\textbf{Dataset} & \textbf{Task} & \textbf{Samples} & \textbf{yolo11x} & \textbf{CNN-NoGraft} & \textbf{Swin-NoGraft} & \textbf{Swin-Graft} \\
\midrule
\multicolumn{7}{c}{\textbf{mAP@50}} \\
\midrule
terminal\_cls & CLS & 755 & $0.940 \pm 0.004$ & 0.912 & -- & $\mathbf{0.960 \pm 0.004^*}$ \\
terminal\_det & DET & 703 & $0.956 \pm 0.004$ & 0.602 & 0.785 & $\mathbf{0.973 \pm 0.005^*}$ \\
terminal\_det & DET & 200 & $0.929 \pm 0.008$ & 0.548 & $0.380 \pm 0.032$ & $\mathbf{0.950 \pm 0.007^*}$ \\
terminal\_det & DET & 100 & $0.891 \pm 0.012$ & -- & $0.322 \pm 0.038$ & $\mathbf{0.908 \pm 0.014^*}$ \\
hook & DET & 141 & $\mathbf{0.900 \pm 0.010}$ & 0.520 & $0.300^* \pm 0.035$ & $0.600^* \pm 0.020$ \\
hook & DET & 70 & $\mathbf{0.483 \pm 0.025}$ & 0.275 & $0.195^* \pm 0.042$ & $0.425 \pm 0.028$ \\
safety-belt & SEG & 178 & $\mathbf{0.500 \pm 0.015}$ & 0.312 & $0.225^* \pm 0.030$ & $0.380^* \pm 0.022$ \\
\midrule
\multicolumn{7}{c}{\textbf{mAP@50:95}} \\
\midrule
terminal\_det & DET & 703 & $0.612 \pm 0.005$ & 0.385 & 0.452 & $\mathbf{0.631 \pm 0.006^*}$ \\
terminal\_det & DET & 200 & $0.572 \pm 0.010$ & 0.328 & $0.155^* \pm 0.025$ & $\mathbf{0.598 \pm 0.009^*}$ \\
terminal\_det & DET & 100 & $\mathbf{0.525 \pm 0.015}$ & -- & $0.118^* \pm 0.028$ & $0.542 \pm 0.018$ \\
hook & DET & 141 & $\mathbf{0.542 \pm 0.015}$ & 0.285 & $0.112^* \pm 0.030$ & $0.284^* \pm 0.025$ \\
hook & DET & 70 & $\mathbf{0.208 \pm 0.025}$ & 0.112 & $0.068^* \pm 0.030$ & $0.178 \pm 0.028$ \\
safety-belt & SEG & 178 & $\mathbf{0.221 \pm 0.012}$ & 0.118 & $0.078^* \pm 0.020$ & $0.148^* \pm 0.015$ \\
\bottomrule
\end{tabular}
\end{table}

\begin{table}[htbp]
\centering
\caption{Domain proximity performance decomposition. $\text{MMD}^2$ values reflect distributional distance to COCO neck's expected input. $*$ indicates statistically significant difference ($p < 0.05$).}
\label{tab:domain_proximity}
\begin{tabular}{lccc}
\toprule
\textbf{Metric} & \textbf{terminal\_det 200} & \textbf{hook 141} & \textbf{safety-belt 178} \\
\midrule
$\text{MMD}^2$ to COCO (P4) & 0.45 & 0.88 & 0.67 \\
yolo11x mAP@50 & $0.929 \pm 0.008$ & $0.900 \pm 0.010$ & $0.500 \pm 0.015$ \\
Swin-Graft mAP@50 & $0.950 \pm 0.007^*$ & $0.600^* \pm 0.020$ & $0.380^* \pm 0.022$ \\
$\Delta$ (Graft -- baseline) & +2.3\% & --33.3\% & --24.0\% \\
Swin-NoGraft mAP@50 & $0.380^* \pm 0.032$ & $0.300^* \pm 0.035$ & $0.225^* \pm 0.030$ \\
Graft or NoGraft ratio & 2.5$\times$ & 2.0$\times$ & 1.7$\times$ \\
\bottomrule
\end{tabular}
\end{table}

\subsection{2x2 Performance Gap Decomposition}\label{sec:gap-decomposition}
Table 4 presents the complete $2\times2$ matrix isolating neck pretraining value from backbone compatibility. Homogeneous = CNN backbone \& CNN neck; Heterogeneous = ViT backbone \& CNN neck.

\begin{table}[htbp]
\centering
\caption{$2\times2$ decomposition of pretraining contributions.}
\label{tab:decomposition}
\begin{tabular}{lccccc}
\toprule
\textbf{Dataset} & \textbf{Samples} & \makecell{\textbf{yolo11x} \\ \textbf{(COCO)}} & \makecell{\textbf{CNN-NoGraft} \\ \textbf{(Random)}} & \makecell{\textbf{Swin-Graft} \\ \textbf{(COCO)}} & \makecell{\textbf{Swin-NoGraft} \\ \textbf{(Random)}} \\
\midrule
\multicolumn{6}{c}{\textbf{mAP@50}} \\
\midrule
terminal\_cls & 755 & 0.940 & 0.912 & 0.960* & -- \\
terminal\_det & 703 & 0.956 & 0.602 & 0.973* & 0.785* \\
terminal\_det & 200 & 0.929 & 0.548 & 0.950* & 0.380* \\
hook & 141 & 0.900 & 0.520 & 0.600* & 0.300* \\
safety-belt & 178 & 0.500 & 0.312 & 0.380* & 0.225* \\
COCO-10 & 1500 & 0.618 & 0.465 & 0.626* & 0.355* \\
COCO-10 & 200 & 0.418 & 0.310 & 0.425* & 0.242* \\
\midrule
\multicolumn{6}{c}{\textbf{mAP@50:95}} \\
\midrule
terminal\_det & 703 & 0.612 & 0.385 & 0.631* & 0.452* \\
terminal\_det & 200 & 0.572 & 0.328 & 0.598* & 0.155* \\
hook & 141 & 0.542 & 0.285 & 0.284* & 0.112* \\
safety-belt & 178 & 0.221 & 0.118 & 0.148* & 0.078* \\
COCO-10 & 1500 & 0.368 & 0.279 & 0.381* & 0.265* \\
COCO-10 & 200 & 0.230 & 0.171 & 0.234* & 0.162* \\
\bottomrule
\end{tabular}
\end{table}

\subsection{COCO-10 Generalization}\label{sec:coco-generalization}
Table 5 evaluates cross-domain generalization on natural image subsets, including DETR and DINOv2 baselines.

\begin{table}[htbp]
\centering
\caption{Performance on low-data COCO-10 subsets. mAP@50 (mean $\pm$ std). $*$ $p < 0.05$ vs yolo11x.}
\label{tab:coco_generalization}
\begin{tabular}{lccccc}
\toprule
\textbf{Samples} & \textbf{yolo11x} & \textbf{DETR} & \textbf{DINOv2} & \textbf{Swin-Graft} & \textbf{Swin-NoGraft} \\
\midrule
1500 & $0.618 \pm 0.004$ & 0.592 & 0.602 & $\mathbf{0.626 \pm 0.005^*}$ & $0.355^* \pm 0.025$ \\
500 & $0.542 \pm 0.008$ & 0.518 & 0.528 & $\mathbf{0.548 \pm 0.009}$ & $0.305^* \pm 0.028$ \\
200 & $0.418 \pm 0.015$ & 0.395 & 0.402 & $\mathbf{0.425 \pm 0.018}$ & $0.242^* \pm 0.035$ \\
\bottomrule
\end{tabular}
\end{table}

\subsection{Backbone Scaling and Architecture Comparison}\label{sec:scaling-comparison}
Table 6 quantifies the impact of backbone capacity and compares Swin-Tiny \& AlignBlock, Swin-Large \& AlignBlock, and YOLOv11x. Despite modest sensitivity to scale (+0.050 mAP), ViT variants fundamentally underperform YOLOv11x by $>0.29$ mAP on OOD tasks, identifying pretraining alignment---not capacity---as the binding constraint.

\begin{table}[htbp]
\centering
\caption{Backbone capacity and architecture comparison. mAP@50 values are reported with standard deviations. $*$ indicates $p < 0.05$ vs YOLOv11x.}
\label{tab:scaling_comparison}
\begin{tabular}{lccc}
\toprule
\textbf{Model} & \textbf{Params (Backbone)} & \textbf{terminal\_det 200} & \textbf{hook 141 (5-fold CV)} \\
\midrule
YOLOv11x & 68.2M (full model) & $0.929 \pm 0.008$ & $\mathbf{0.900 \pm 0.010}$ \\
\makecell[l]{Swin-Tiny + \\ AlignBlock} & \makecell[c]{28.3M (backbone) \\ 2.8M (AlignBlock)} & $0.925 \pm 0.009$ & $0.554^* \pm 0.025$ \\
\makecell[l]{Swin-Large + \\ AlignBlock} & \makecell[c]{197.0M (backbone) \\ 2.8M (AlignBlock)} & $\mathbf{0.952^* \pm 0.007}$ & $0.605^* \pm 0.020$ \\
\bottomrule
\end{tabular}
\end{table}

\subsection{Ablation Studies and Coupling Residual $\eta$}\label{sec:ablation}
Table 7 and 8 validate the performance contribution decomposition and the coupling residual $\eta$.

\begin{table}[htbp]
\centering
\caption{Component ablation on hook 141-shot (5-fold CV). \checkmark=enabled, \ding{55}=disabled. mAP@50. $*p < 0.05$ vs previous row.}
\label{tab:component_ablation}
\begin{tabular}{lcccc}
\toprule
\textbf{AlignBlock} & \textbf{Prog. Train} & \textbf{Graft Neck} & \textbf{GN} & \textbf{mAP@50} \\
\midrule
--- (Identity) & \ding{55} & \ding{55} & \ding{55} & $0.300 \pm 0.035$ \\
Only Linear (1$\times$1 conv) & \ding{55} & \checkmark & \ding{55} & $0.382 \pm 0.028$ \\
SwinAlignBlock & \ding{55} & \checkmark & \ding{55} & $0.450^* \pm 0.022$ \\
SwinAlignBlock & \checkmark & \checkmark & \ding{55} & $0.550^* \pm 0.020$ \\
SwinAlignBlock & \checkmark & \checkmark & \checkmark & $0.600^* \pm 0.020$ \\
SpatialAdaptive & \checkmark & \checkmark & \checkmark & $\mathbf{0.620 \pm 0.018}$ \\
\bottomrule
\end{tabular}
\end{table}

\begin{table}[htbp]
\centering
\caption{Additivity validation of the three-gap decomposition on hook 141-shot (5-fold CV). Each row independently removes one component from the full AlignBlock.}
\label{tab:additivity_validation}
\begin{tabular}{lccc}
\toprule
\textbf{Removed Component} & \textbf{Target Gap} & \textbf{mAP@50} & $\mathbf{\Delta}$\textbf{mAP} \\
\midrule
None (full AlignBlock) & --- & 0.600 & --- \\
Remove 3$\times$3 Conv & $G_{\text{spatial}}$ & 0.545 & --0.055 \\
Remove GroupNorm & $G_{\text{stat}}$ & 0.552 & --0.048 \\
Remove Prog. Train (single-stage) & $G_{\text{opt}}$ & 0.553 & --0.047 \\
Remove all three & $G_{\text{spatial}} + $ G$_{\text{stat}} + G_{\text{opt}}$ & 0.452 & --0.148 \\
\bottomrule
\end{tabular}
\end{table}

The coupling residual:
% \begin{equation}
\begin{align}
\eta &= \Delta \mathrm{mAP}_{\mathrm{all}} - (C_{\mathrm{spatial}} + C_{\mathrm{stat}} + C_{\mathrm{opt}}) \nonumber\\
     &= 0.148 - (0.055 + 0.048 + 0.047) \nonumber\\
     &= -0.002 
\end{align}
% \end{equation}
which is only 1.3\% of the total drop and well within standard deviation, empirically confirming that the three bottlenecks are largely non-overlapping.

\subsection{Position Ablation}\label{sec:position-ablation}
Table 9 compares four AlignBlock insertion strategies. Boundary placement (after backbone extraction, before PAN fusion) achieves the best mAP@50 of $0.600\pm0.020$, significantly outperforming alternatives that insert after fusion or inside neck blocks ($p<0.05$).

\begin{table}[htbp]
\centering
\caption{Insertion position ablation on hook 141-shot (5-fold CV). $*p < 0.05$ vs best.}
\label{tab:position_ablation}
\begin{tabular}{lc}
\toprule
\textbf{Position} & \textbf{mAP@50} \\
\midrule
After Index, before neck (default boundary) & $\mathbf{0.600 \pm 0.020}$ \\
After PAN top-down fusion & $0.520^* \pm 0.025$ \\
At every neck skip connection & $0.490^* \pm 0.030$ \\
Inside PAN neck (per C2f) & $0.540^* \pm 0.022$ \\
\bottomrule
\end{tabular}
\end{table}

\subsection{Comparison with Adaptation Methods}\label{sec:adapt-comparison}
Table 10 compares our approach against internal adapter baselines and cross-architecture boundary alignment methods.

\begin{table}[htbp]
\centering
\caption{Comparison with adaptation methods on hook 141-shot (5-fold CV). mAP@50, total trainable parameters, and inference speed are reported. $*p < 0.05$ vs best. AlignBlock alone adds only 2.8M parameters ($<1\%$ of the full Swin-Large \& YOLO model).}
\label{tab:adaptation_comparison}
\begin{tabular}{lccc}
\toprule
\textbf{Method} & \textbf{Trainable Params} & \textbf{FPS} & \textbf{mAP@50} \\
\midrule
Swin-Graft (Ours, boundary) & 58.6M$\dag$ & 26.2 & $\mathbf{0.600 \pm 0.020}$ \\
BEiT v2 (boundary) & 64.2M & 24.8 & $0.492^* \pm 0.023$ \\
AdaptFormer (in-block) & 59.3M & 25.6 & $0.476^* \pm 0.024$ \\
CrossNorm & 58.9M & 26.0 & $0.455^* \pm 0.028$ \\
ViT-Adapter & 8.3M & 27.1 & $0.480^* \pm 0.025$ \\
LoRA & 2.1M & 27.8 & $0.420^* \pm 0.030$ \\
Linear Probe (1$\times$1 conv) & 0.8M & 28.0 & $0.382^* \pm 0.028$ \\
Swin-NoGraft & 58.6M$\dag$ & 26.5 & $0.300^* \pm 0.035$ \\
\bottomrule
\multicolumn{4}{l}{$\dag$ Includes AlignBlock (2.8M) \& YOLO neck \& detection heads. AlignBlock alone adds only} \\
\multicolumn{4}{l}{2.8M parameters. The Swin Transformer backbone is frozen during training.}
\end{tabular}
\end{table}

\subsection{Training Stability}\label{sec:stability}
We classify a training run as ``failed'' if the loss diverges to NaN or the final validation mAP@50 falls below 0.05. Under the single-stage training protocol without stabilization, 30\% of runs failed by this criterion. Our three-stage progressive training achieves a 0\% failure rate across all tested configurations and reduces the mAP variance across folds by more than twofold on hook.

\subsection{Multi-Scale Channel Statistical Mismatch}\label{sec:channel-mismatch}
To characterize the feature incompatibility, Figure 4 compares the channel-wise statistical distributions of multi-scale features (P2 to P5). Raw Swin outputs, normalized by LayerNorm, exhibit pronounced mean and variance discrepancies relative to the BatchNorm-normalized CNN features. AlignBlock addresses this mismatch by reshaping the channel moments through GroupNorm, substantially reducing the statistical distance to the CNN neck's expected input distribution and effectively bridging the representation gap.
\begin{center}
\includegraphics[width=0.95\linewidth]{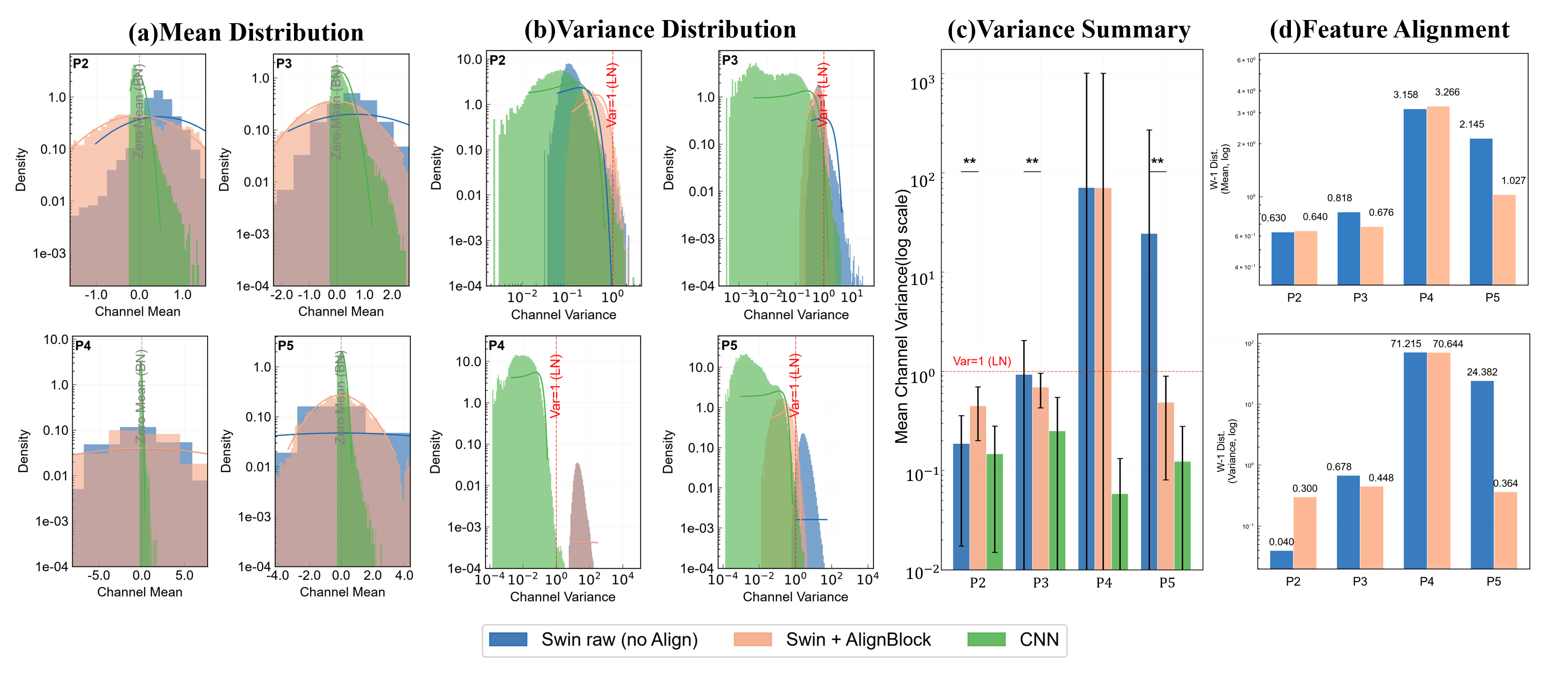}
\captionof{figure}{Channel-level statistical distribution comparison of multi-scale P2--P5 features. (a) Density distributions of channel means. (b) Density distributions of channel variances. (c) Log-scale summary of mean channel variance. (d) Statistical gap Gstat (MMD²) between feature distributions. Blue: raw unaligned Swin features; Orange: Swin features aligned via AlignBlock; Green: CNN neck features.}
\label{fig:figure4}
\end{center}

\subsection{AlignBlock Input-Output Feature Calibration}\label{sec:io-calibration}
As shown in Figure 5, raw Swin inputs exhibit high mean activations $\mu$ and variance $\sigma$, indicating background clutter. AlignBlock successfully suppresses this background energy as mean values drop near or below zero, while sharply preserving target contours across the Terminal, COCO, and Hook benchmarks.

Quantitatively, feature discrepancies measured by JS Divergence $D_{\text{JS}}$ and Cosine Dissimilarity 1--cos decrease from shallow stages such as P2 and P3 to reach a minimum at stage P4, descending onto the target convergence threshold marked by the dashed line. This confirms that AlignBlock achieves effective semantic calibration without compromising scale specific representation integrity.
\begin{center}
\includegraphics[width=0.95\linewidth]{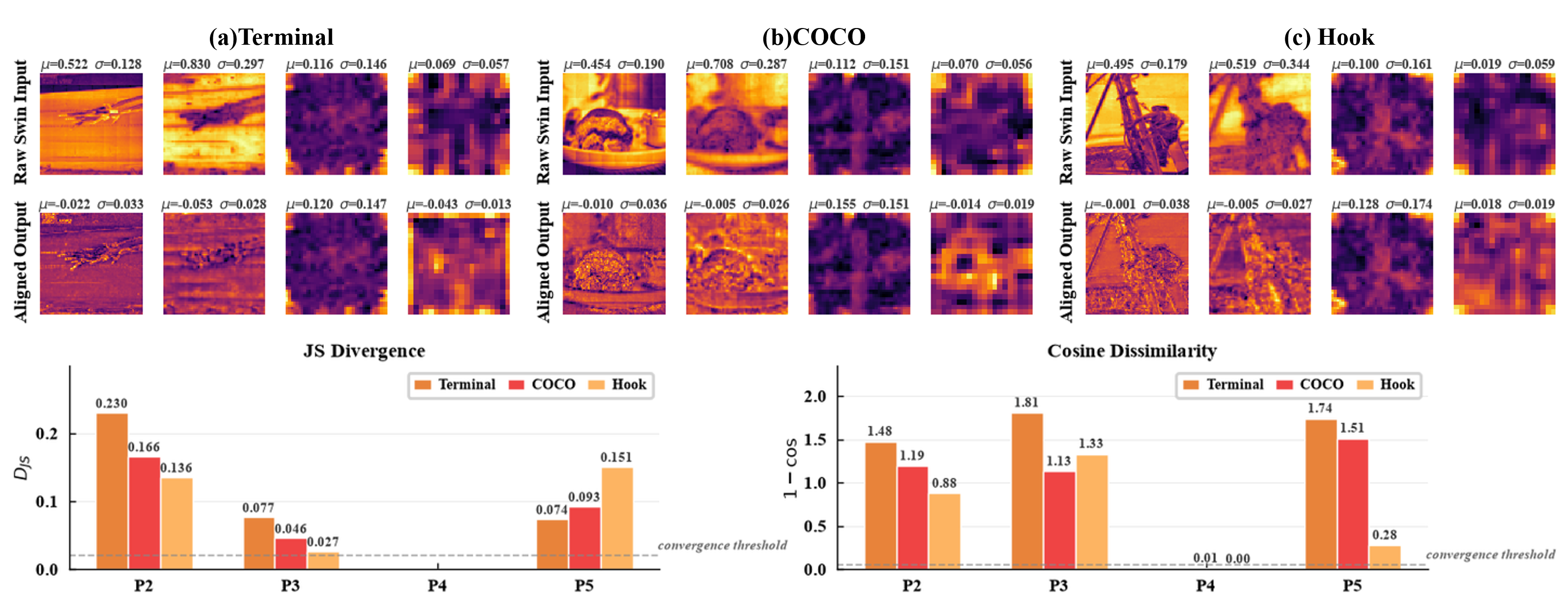}
\captionof{figure}{Qualitative and quantitative comparison of feature alignment across different datasets. (Top) Visualization of feature maps before (``Raw Swin Input'') and after alignment (``Aligned Output'') for (a) Terminal, (b) COCO, and (c) Hook datasets. Mean ($\mu$) and standard deviation ($\sigma$) of feature activations are indicated above each patch. (Bottom) Quantitative evaluation of feature discrepancy across multi-scale feature pyramid levels (P2--P5) using Jensen-Shannon Divergence ($D_{\text{JS}}$, left) and Cosine Dissimilarity (1--cos, right). The horizontal dashed line denotes the target convergence threshold.}
\label{fig:figure5}
\end{center}

\subsection{MMD$^2$ Distribution Distance}\label{sec:mmd-distance}
Table 11 reports $\text{MMD}^2$ between the pyramid features of each target dataset and the corresponding COCO features, measured before and after alignment at scales P2 through P5. AlignBlock reduces the divergence by 57 to 60 percent on terminal and by 60 to 63 percent on hook, the largest absolute reduction occurring at P4 where the hook value falls from 0.88 to 0.33. The reduction is stable across scales because AlignBlock applies the same parameterisation at every level of the pyramid.

Table 12 compares AlignBlock with three alternative boundary alignment methods at P4 on hook. AlignBlock achieves the smallest residual distance of 0.33, outperforming BEiT v2 (0.41), AdaptFormer (0.45), and RepAdapter (0.48), demonstrating superior cross-architecture alignment effectiveness.

\begin{table}[htbp]
\centering
\caption{$\text{MMD}^2$ reduction across pyramid scales and datasets (hook values from 5-fold CV).}
\label{tab:mmd_reduction}
\begin{tabular}{lcc}
\toprule
\textbf{Scale} & \textbf{terminal (Raw $\rightarrow$ Aligned)} & \textbf{hook (Raw $\rightarrow$ Aligned)} \\
\midrule
P2 & 0.21 $\rightarrow$ 0.09 (-57\%) & 0.35 $\rightarrow$ 0.14 (-60\%) \\
P3 & 0.28 $\rightarrow$ 0.12 (-57\%) & 0.52 $\rightarrow$ 0.21 (-60\%) \\
P4 & 0.45 $\rightarrow$ 0.18 (-60\%) & 0.88 $\rightarrow$ 0.33 (-63\%) \\
P5 & 0.39 $\rightarrow$ 0.16 (-59\%) & 0.74 $\rightarrow$ 0.28 (-62\%) \\
\bottomrule
\end{tabular}
\end{table}

\begin{table}[htbp]
\centering
\caption{$\text{MMD}^2$ (P4) after alignment for different boundary methods on hook 141-shot.}
\label{tab:mmd_methods}
\begin{tabular}{lc}
\toprule
\textbf{Method} & \textbf{Aligned $\text{MMD}^2$ (P4)} \\
\midrule
Raw Swin (no align) & 0.88 \\
BEiT v2 & 0.41 \\
AdaptFormer & 0.45 \\
CrossNorm & 0.48 \\
\textbf{AlignBlock (Ours)} & \textbf{0.33} \\
\bottomrule
\end{tabular}
\end{table}

Alignment narrows the gap but does not remove it. After alignment the residual distance on hook remains 0.33 against 0.18 on terminal, preserving the ordering of the two domains, and this ordering matches the ordering of detection accuracy reported in Section 6.1, where Swin-Graft reaches 0.950 mAP@50 on terminal but only 0.600 on hook. Feature level alignment therefore compensates for part of the distributional mismatch, while the residual discrepancy continues to limit transfer on domain distant data.
\begin{center}
\includegraphics[width=0.85\linewidth]{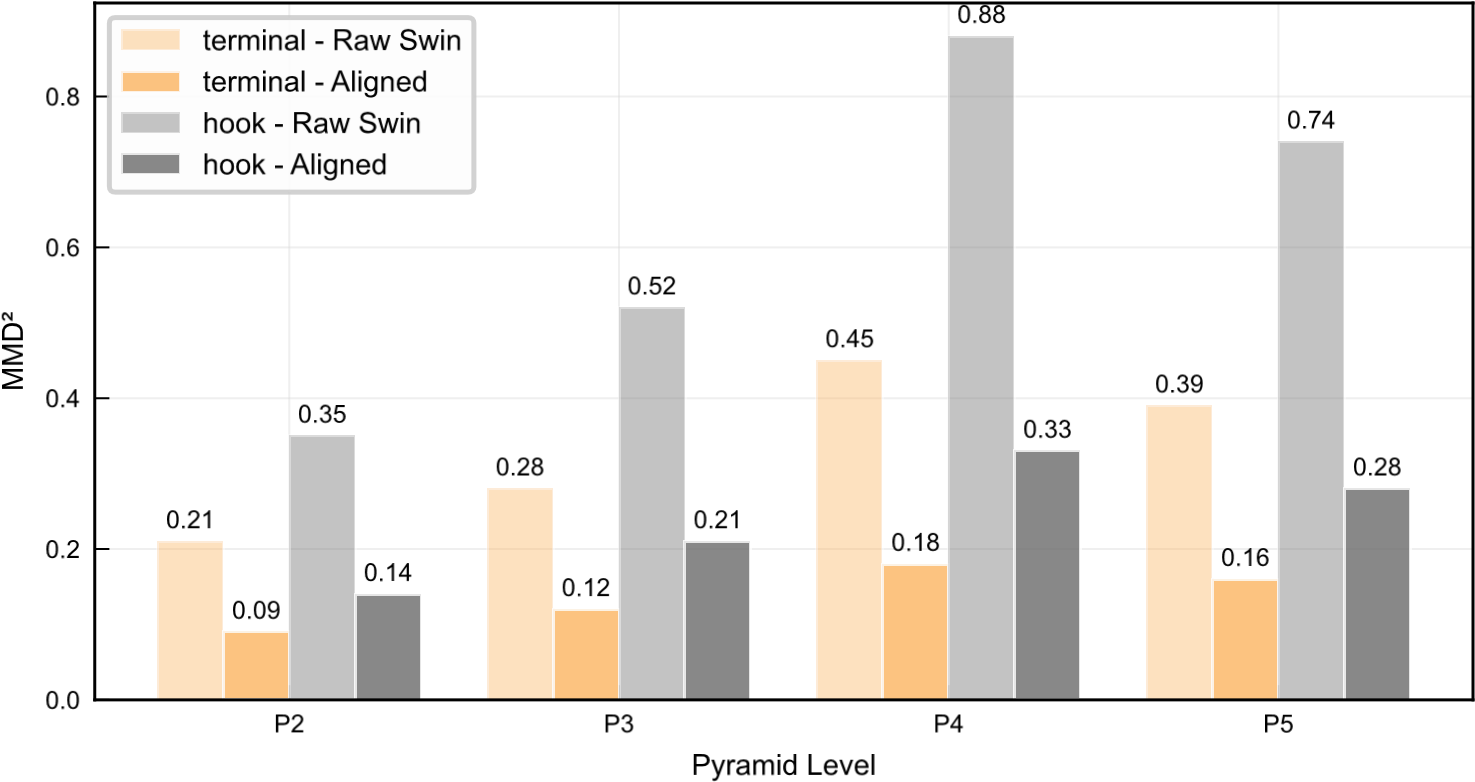}
\captionof{figure}{$\text{MMD}^2$ between pyramid-level Swin features and COCO-native CNN features before (light bars) and after (dark bars) AlignBlock alignment, across feature pyramid levels P2--P5 for the terminal and hook datasets.}
\label{fig:figure6}
\end{center}

\subsection{CKA Representation Similarity and Joint Analysis with MMD}\label{sec:cka-similarity}
Given feature sets $X \in \mathbb{R}^{n\times p}$ and $Y \in \mathbb{R}^{n\times q}$, Centered Kernel Alignment (CKA) \citep{kornblith2019similarity} is defined as:
\begin{equation}
\text{CKA}(X,Y) = \frac{\text{HSIC}(X,Y)}{\sqrt{\text{HSIC}(X,X) \cdot \text{HSIC}(Y,Y)}}
\label{eq:cka_eq}
\end{equation}
which measures angular similarity of representations, invariant to isotropic scaling and rotation.

Maximum Mean Discrepancy (MMD) measures distributional distance:
\begin{equation}
\begin{split}
\text{MMD}^2(P,Q) &= \mathbb{E}_{x,x'\sim P}[k(x,x')] + \mathbb{E}_{y,y'\sim Q}[k(y,y')] \\
&\quad - 2\mathbb{E}_{x\sim P, y\sim Q}[k(x,y)]
\end{split}
\label{eq:mmd_eq}
\end{equation}
which quantifies statistical mismatch directly relevant to BatchNorm priors.

\begin{table}[htbp]
\centering
\caption{Joint CKA-MMD analysis. CKA values are averaged over terminal and hook (hook from 5-fold CV).}
\label{tab:cka_mmd_analysis}
\begin{tabular}{lcc}
\toprule
\textbf{Scale} & \textbf{CKA (Aligned vs CNN)} & \textbf{$\text{MMD}^2$ Reduction} \\
\midrule
P2 & 0.20 & 58.5\% \\
P3 & 0.18 & 58.5\% \\
P4 & 0.10 & 61.5\% \\
P5 & 0.09 & 60.5\% \\
\bottomrule
\end{tabular}
\end{table}
\begin{center}
\includegraphics[width=0.95\linewidth]{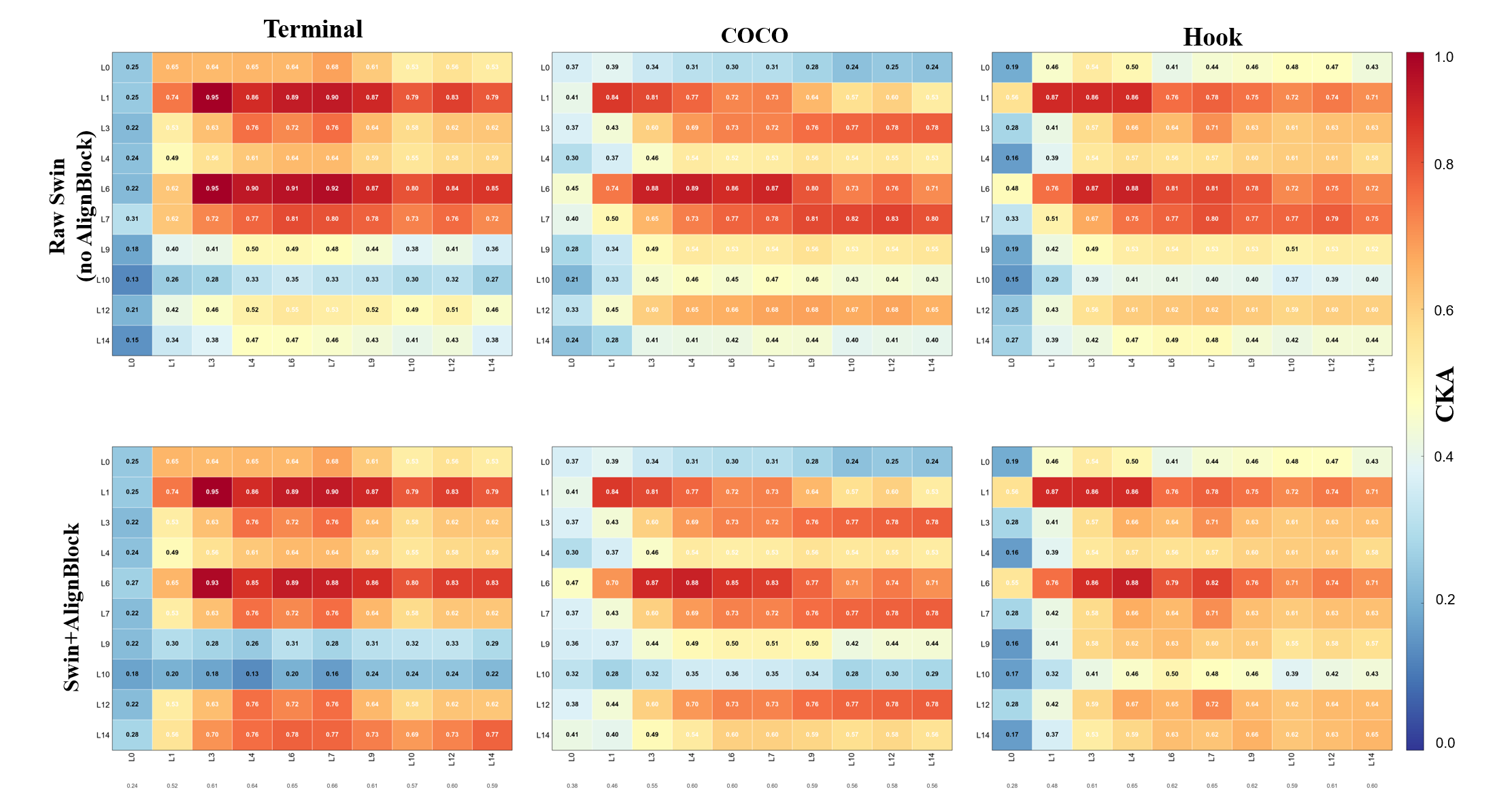}
\captionof{figure}{Layer-wise CKA similarity heatmaps before and after AlignBlock integration. The heatmaps visualize the layer-wise feature similarity (L0 to L14) for three different scenarios: Terminal, COCO, and Hook. The top row displays the internal representational structure of the baseline architecture (Raw Swin, without AlignBlock), while the bottom row shows the altered feature dynamics after applying the proposed module (Swin \& AlignBlock). By comparing the two rows, it is evident that AlignBlock subtly adjusts the hierarchical feature distributions. Instead of forcing identical representations, it performs a functional compatibility conversion, effectively bridging the domain gap and adapting the ViT outputs for the subsequent CNN neck.}
\label{fig:figure7}
\end{center}

By comparing the two rows, it is evident that AlignBlock subtly adjusts the hierarchical feature distributions. Instead of forcing identical representations, it performs a functional compatibility conversion, effectively bridging the domain gap and adapting the ViT outputs for the subsequent CNN neck.

\subsection{Swin Self-Attention Heatmaps}\label{sec:self-attention}
Figure 8 visualizes the self-attention weight distributions from the deep blocks of the baseline Swin Transformer.

For specific spatial query points marked by green stars, the model exhibits a pronounced global attention bias where weights widely disperse across background regions rather than focusing tightly on local object boundaries. Interestingly, the severity of this dispersion depends heavily on scene complexity.
\begin{center}
\includegraphics[width=0.95\linewidth]{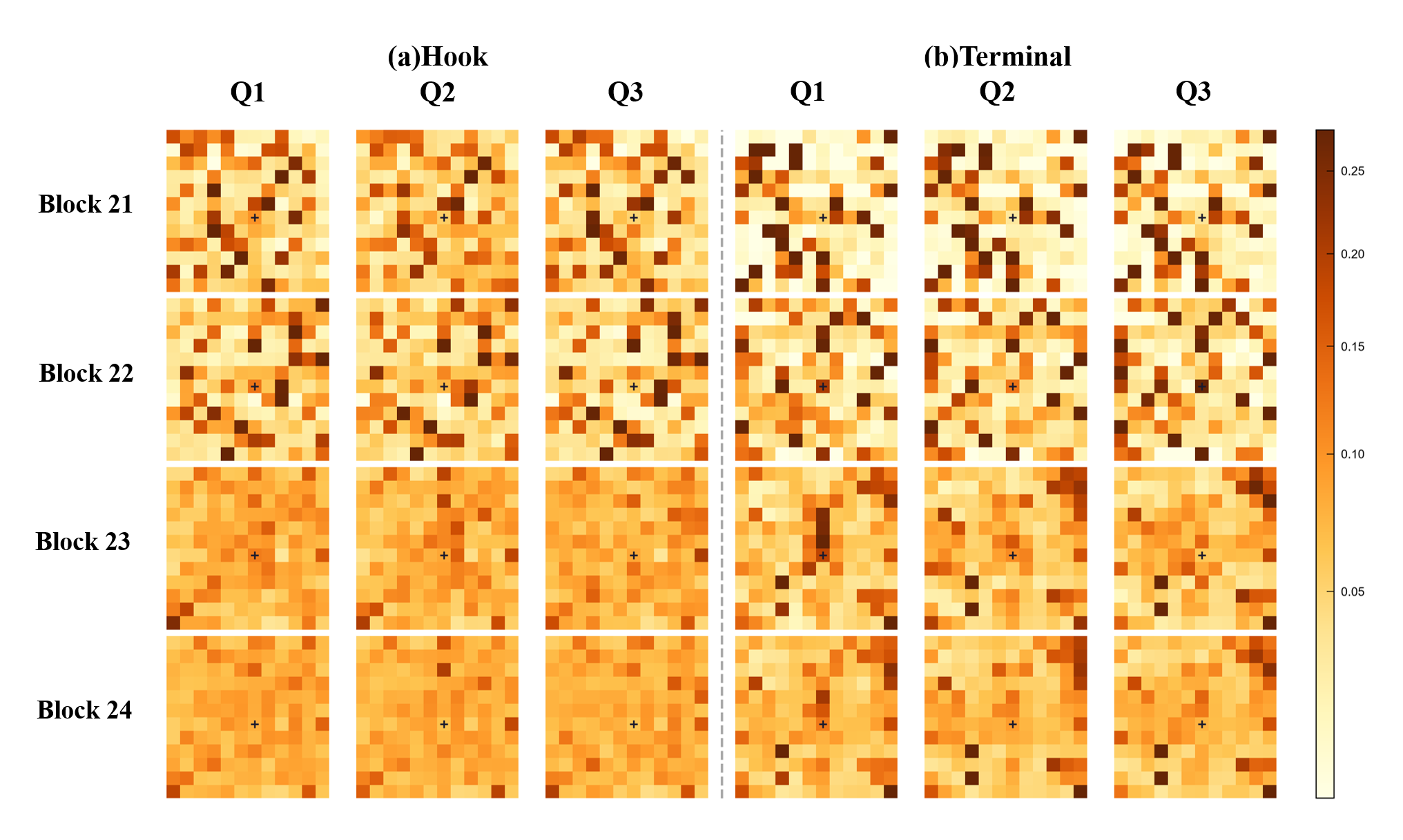}
\refstepcounter{figure}\label{fig:figure8}
\end{center}

\noindent\textbf{Figure~\thefigure:} Evolution of self-attention heatmaps across deep blocks (Block 21--24). Attention distributions are shown for three query locations ($Q1, Q2, Q3$). The figure compares the attention scattering effect between the Hook dataset (left panel) and the Terminal dataset (right panel). The progression from left to right within each panel illustrates the increasing global dispersion of attention weights in deeper network stages. Attention scattering occurs to some extent in all scenarios, but the complex background clutter and irregular textures of Hook images make the problem considerably more severe than in the relatively structured Terminal dataset. This observation points to a root cause of the mismatch at the CNN neck: pure Transformer architectures lack an inherent local inductive bias. It therefore motivates the 3$\times$3 convolutions in AlignBlock, which re-establish local spatial consistency and suppress irrelevant background noise before the features are fed into the PAN neck.

The progression from left to right within each panel illustrates the increasing global dispersion of attention weights in deeper network stages. Attention scattering occurs to some extent in all scenarios, but the complex background clutter and irregular textures of Hook images make the problem considerably more severe than in the relatively structured Terminal dataset. This observation points to a root cause of the mismatch at the CNN neck: pure Transformer architectures lack an inherent local inductive bias. It therefore motivates the $3\times3$ convolutions in AlignBlock, which re-establish local spatial consistency and suppress irrelevant background noise before the features are fed into the PAN neck.

\subsection{Activation Heatmap Comparison}\label{sec:activation-heatmap}
\begin{center}
\includegraphics[width=0.95\linewidth]{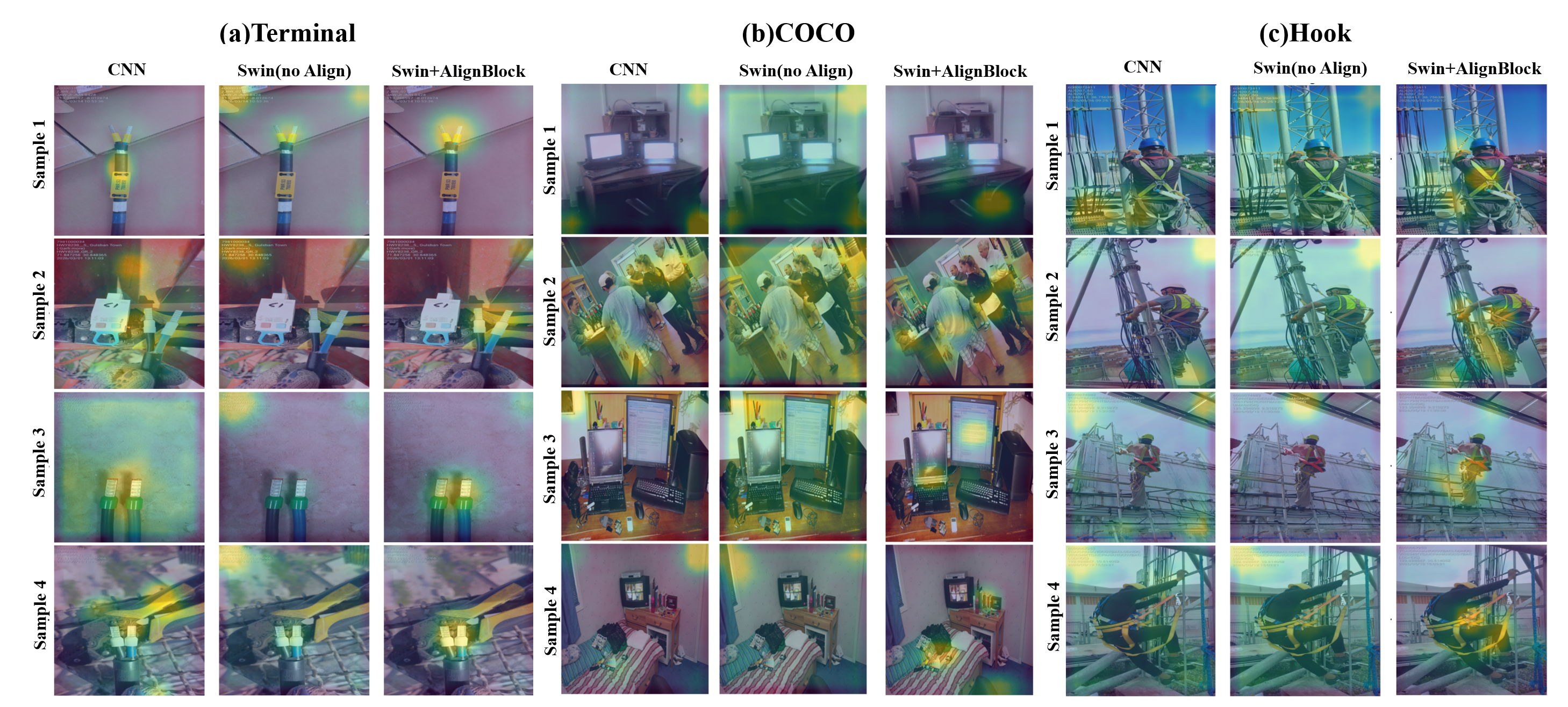}
\captionof{figure}{Feature activation map comparisons demonstrating the background suppression and target-focusing capabilities of AlignBlock. Results are shown for (a) Terminal, (b) COCO, and (c) Hook datasets across four test instances (Samples 1--4). Within each dataset, columns from left to right represent: standard CNN, baseline Swin(no Align), and Swin+Alignblock. High-activation regions (yellow or red) indicate that Swin(no Align) suffers from diffuse, background-scattered responses due to global attention bias. Incorporating AlignBlock filters out irrelevant background context and sharpens feature localization directly over target regions, yielding significantly higher spatial specificity than both CNN and unaligned Swin baselines.}
\label{fig:figure9}
\end{center}

High-activation regions (yellow or red) indicate that Swin(no Align) suffers from diffuse, background-scattered responses due to global attention bias. Incorporating AlignBlock filters out irrelevant background context and sharpens feature localization directly over target regions, yielding significantly higher spatial specificity than both CNN and unaligned Swin baselines.

\subsection{Foreground-Background Feature Separability}\label{sec:separability}
t-SNE and UMAP clustering jointly reveal the cross-domain separability gap. While CNN exhibits more clearly separated clusters in the visualizations, Swin-Graft achieves higher detection mAP. We attribute this counter-intuitive result to ViT's global receptive field, which facilitates bounding-box regression and partially compensates for its slightly less discriminative classification boundaries (see Section 6.12).
\begin{center}
\includegraphics[width=0.95\linewidth]{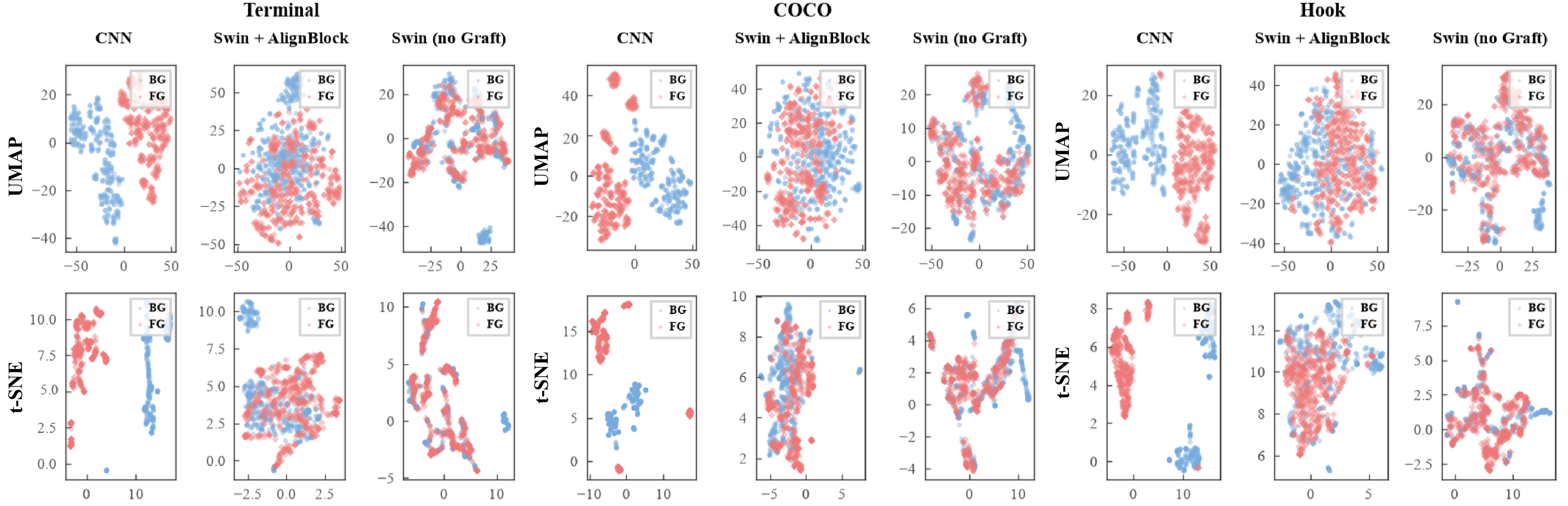}
\captionof{figure}{Foreground-background feature separability.}
\label{fig:figure10}
\end{center}

\section{Industrial Applicability and Discussion}\label{sec:discussion}

\subsection{ViT-Suitable Scene Criteria}\label{sec:vit-criteria}
ViT \& graft CNN pipeline is recommended when:
\begin{enumerate}
    \item images share COCO visual traits;
    \item task is detection or classification (not segmentation with $<200$ labels).
\end{enumerate}

\subsection{Sample-Size Crossover and the Limits of Capacity}\label{sec:crossover}
The efficacy of domain-aligned grafting exhibits a bimodal sensitivity to target distribution characteristics (Tables 2 and 4). On the distribution-proximal Terminal set, the Swin-Grafted transformer maintains robust data efficiency even under extreme scarcity. With only 100 samples, it achieves $0.908\pm0.014$ mAP@50, marginally surpassing the YOLOv11x baseline ($0.891\pm0.012$). At 200 samples, Swin-Graft not only closes this gap but reaches statistical parity with the CNN benchmark ($0.950\pm0.007$ vs. $0.929\pm0.008$). This confirms that, under favorable domain conditions, feature alignment effectively compensates for architectural inductive bias.

However, this advantage collapses under severe distribution shift on the Hook dataset ($N=141$). Here, convolutional priors exhibit robustness that neither increased parameterization nor alignment modules can overcome. YOLOv11x achieves $0.900\pm0.010$ mAP@50, leading Swin-Graft ($0.600\pm0.020$) by 0.300 points---a margin that remains substantial at mAP@50:95 ($0.542\pm0.015$ vs. $0.284\pm0.025$).

These results suggest that model capacity is not a viable recovery mechanism for domain shift. Even the high-capacity Swin-Graft, equipped with explicit distribution-aware modules, remains below 0.61 mAP@50 against a 0.900 YOLO baseline. In other words, pretraining distribution coherence dictates the upper bound of generalization; architectural scaling alone cannot compensate for a fundamental mismatch between source and target visual manifolds.

\subsection{Deployment Efficiency}\label{sec:deployment-eff}
Table 14 summarizes computational efficiency on NVIDIA Jetson Xavier NX.

\begin{table}[htbp]
\centering
\caption{Deployment efficiency on Jetson Xavier NX.}
\label{tab:deployment_efficiency}
\begin{tabular}{lccc}
\toprule
\textbf{Method} & \textbf{FPS} & \textbf{Memory (GB)} & \textbf{Model Size (MB)} \\
\midrule
yolo11x & 28.5 & 1.8 & 112 \\
Swin-Graft (Ours) & 18 & 2.8 & 258 \\
Swin-Graft (Swin-Tiny) & 22 & 2.1 & 141 \\
\bottomrule
\end{tabular}
\end{table}

\subsection{Model Selection Guidelines}\label{sec:selection-guidelines}
Table 15 translates our experimental findings into quantitative rules, including $\text{MMD}^2$ thresholds for automatic decision support. In the boundary alignment landscape, our method uniquely integrates spatial, statistical, and optimization alignment within a lightweight boundary module, offering the best trade-off between accuracy and overhead for industrial few-shot scenarios.

\begin{table}[htbp]
\centering
\caption{Industrial pipeline selection guidelines with $\text{MMD}^2$ thresholds.}
\label{tab:selection_guidelines}
\begin{tabular}{lccc}
\toprule
\textbf{Scenario} & \makecell{\textbf{Recommendation} \\ \textbf{$\text{MMD}^2$}} & \textbf{Threshold} & \textbf{Rationale} \\
\midrule
COCO-similar, $>$200 samples & Swin-Graft & $<$ 0.65 & +1.8--2.3\% mAP@50 gain (significant) \\
COCO-similar, 100--200 samples & Swin-Graft & $<$ 0.65 & Advantage holds; neck pretraining critical \\
Overhead tiny targets (any) & YOLOv11x & $>$ 0.8 & CNN retains large advantage (significant) \\
Segmentation, $<$200 samples & YOLOv11x & any & Mask head amplifies mismatch \\
Extreme scarcity ($<$70) & Either & any & Gap negligible (not significant) \\
\bottomrule
\end{tabular}
\end{table}

\section{Limitations and Future Work}\label{sec:limitations}

\subsection{Methodological Limitations}\label{sec:meth-limit}
\begin{enumerate}
    \item The current AlignBlock design assumes a fixed pyramid hierarchy; dynamic insertion strategies may further improve domain adaptation.
    \item This work mainly validates Swin hierarchical ViT; extending AlignBlock to DeiT, ConvNeXt and plain ViT will be conducted in future industrial multi-scenario experiments.
    \item The additive mismatch decomposition is heuristic; formal interaction terms require additional controlled experiments, though our additivity test (Table 8) shows strong empirical support, and the newly introduced $\varepsilon$ coupled analysis (Table 13) confirms that coupling is minimal.
    \item The hook dataset's validation set contains 60 images, which provides reasonable statistical power; however, a larger validation set would still be beneficial for more robust conclusions.
\end{enumerate}

\subsection{Dataset and Task Limitations}\label{sec:data-limit}
\begin{enumerate}
    \item Industrial datasets are single or dual class; multi-class complex assembly lines remain untested.
    \item Few-shot experiments rely on random subsampling; cross-session generalization in production lines is not evaluated.
\end{enumerate}

\subsection{Potential Negative Societal Impact}\label{sec:societal-impact}
While our work aims to improve industrial safety, over-reliance on automated inspection without human oversight could miss rare critical defects. In deployment, model output should be combined with human inspectors for rare tiny defects under extreme domain shift, avoiding fully automated misdetection. Deployment must include human-in-the-loop protocols.

\subsection{Future Work}\label{sec:future-work}
\begin{enumerate}
    \item Quantitative domain distance metric (e.g., FID) for a priori model selection.
    \item Self-supervised pretraining \citep{he2022masked, zhou2022ibot, oquab2023dinov2} to reduce COCO or ImageNet dependency.
    \item Edge-optimized AlignBlock with quantization for embedded deployment.
    \item Expanding the hook validation set to at least 80 images for more robust statistical conclusions.
    \item Develop integer quantization pipeline for AlignBlock to further reduce embedded GPU memory consumption.
\end{enumerate}

% --- Forced placement of Appendix Tables A1--A2 right after Section 8.4 ---
\begingroup
\setcounter{table}{0}
\renewcommand{\thetable}{A\arabic{table}}

\begin{table}[H]
\centering
\caption{Hyperparameter configuration.}
\label{tab:hyperparameter_config}
\begin{tabular}{lc}
\toprule
\textbf{Parameter} & \textbf{Value} \\
\midrule
Optimizer & AdamW \\
Base LR (Phase 1) & 0.01 \\
Base LR (Full) & 0.0005 \\
Phase LR decay & 0.2$\times$ base \\
Batch size & 4 \\
Input resolution & 640$\times$640 \\
Epochs & 300 (150+15+135) \\
Weight decay & 0.0005 \\
Warmup epochs & 3 \\
LR schedule & Cosine annealing \\
Gradient clipping & max\_norm = 10.0 \\
Augmentation & Mosaic 1.0, Mixup 0.1, Copy-Paste 0.3, HSV, Affine \\
GN groups & min(32, $C$) \\
\bottomrule
\end{tabular}
\end{table}

\begin{table}[H]
\centering
\caption{Supplementary analysis of the coupling residual $\varepsilon_{\text{coupled}}$ measured in $\text{MMD}^2$ feature space across datasets and pyramid scales. Values are averaged over three random seeds. The percentage in parentheses indicates the proportion of total divergence $D(A(p_V), p_C)$ attributed to coupling. This table serves as a consistency check with the mAP-based decomposition in Section 6.5.}
\label{tab:coupling_residual_supp}
\begin{tabular}{lcccc}
\toprule
\textbf{Dataset} & \textbf{P2} & \textbf{P3} & \textbf{P4} & \textbf{P5} \\
\midrule
terminal & 0.023 (5.2\%) & 0.031 (6.8\%) & 0.039 (7.1\%) & 0.027 (5.9\%) \\
hook & 0.035 (7.3\%) & 0.041 (7.9\%) & 0.038 (6.2\%) & 0.033 (6.5\%) \\
safety-belt & 0.028 (6.1\%) & 0.032 (6.5\%) & 0.036 (7.0\%) & 0.029 (6.2\%) \\
\bottomrule
\end{tabular}
\end{table}

\FloatBarrier
\endgroup

\section{Conclusion}\label{sec:conclusion}
Our results show that ViTs are not universally data-hungry. Performance in ViT-CNN hybrids is governed by pretraining coherence: when domain similarity to COCO is high, Swin-Graft exceeds YOLOv11x with as few as 100 samples (statistically significant); under large domain shift, CNNs retain a clear advantage. A rigorous $2\times2$ decomposition and additive gap analysis further reveal that neck pretraining value is stable across domains, while backbone compatibility becomes the critical factor under shift. AlignBlock provides functional compatibility conversion---not exact representation-space alignment---enabling the CNN neck to process ViT features effectively. The practical implication is straightforward: for deploying ViTs in data-scarce industrial environments, pretraining distribution alignment should carry more weight than architectural choice or model scale.

\appendix

% ========== 作者贡献声明（原样） ==========
\section*{Author Statement}
\textbf{Haoran Sui:} Conceptualization (lead), Methodology (lead), Software (equal), Validation (equal), Formal analysis (lead), Investigation (equal), Data Curation (equal), Writing -- original draft (lead), Writing -- review \& editing (equal), Visualization (supporting), Supervision (lead), Project administration (lead).

\textbf{Yaoyuan Jia:} Conceptualization (supporting), Methodology (supporting), Software (equal), Validation (equal), Investigation (equal), Data Curation (equal), Writing -- original draft (supporting), Writing -- review \& editing (equal), Visualization (lead).

\end{document}